 \documentclass[pmlr,twocolumn,10pt]{jmlr} 

\usepackage{booktabs}
\usepackage{siunitx}
\usepackage{multirow}

\theorembodyfont{\upshape}
\theoremheaderfont{\scshape}
\theorempostheader{:}
\theoremsep{\newline}

\jmlrvolume{225}
\jmlryear{2023}
\jmlrsubmitted{225}
\jmlrpublished{225}
\jmlrworkshop{Machine Learning for Health (ML4H) 2023} 

 \title[Multimodal Pretraining of Medical Time Series and Notes]{Multimodal Pretraining of Medical Time Series and Notes}

\author{%
\Name{Ryan King} \Email{kingrc15@tamu.edu}\\
\addr Department of Computer Science \& Engineering, Texas A\&M University, United States
\AND
\Name{Tianbao Yang}\footnotemark[1] \Email{tianbao\-yang@tamu.edu}\\
\addr Department of Computer Science \& Engineering, Texas A\&M University, United States
\AND
\Name{Bobak J. Mortazavi}\footnotemark[1] \Email{bobakm@tamu.edu}\\
\addr Department of Computer Science \& Engineering, Texas A\&M University, United States\\
*Equal contribution senior author
}

\begin{document}

\maketitle

\begin{abstract}
Within the intensive care unit (ICU), a wealth of patient data, including clinical measurements and clinical notes, is readily available. This data is a valuable resource for comprehending patient health and informing medical decisions, but it also contains many challenges in analysis. Deep learning models show promise in extracting meaningful patterns, but they require extensive labeled data, a challenge in critical care. To address this, we propose a novel approach employing self-supervised pretraining, focusing on the alignment of clinical measurements and notes. Our approach combines contrastive and masked token prediction tasks during pretraining. Semi-supervised experiments on the MIMIC-III dataset demonstrate the effectiveness of our self-supervised pretraining. In downstream tasks, including in-hospital mortality prediction and phenotyping, our pretrained model outperforms baselines in settings where only a fraction of the data is labeled, emphasizing its ability to enhance ICU data analysis. Notably, our method excels in situations where very few labels are available, as evidenced by an increase in the AUC-ROC for in-hospital mortality by 0.17 and in AUC-PR for phenotyping by 0.1 when only 1\% of labels are accessible. This work advances self-supervised learning in the healthcare domain, optimizing clinical insights from abundant yet challenging ICU data.
\end{abstract}


\section{Introduction}
\label{sec:instructions}

In intensive care units (ICUs), the admission of a patient triggers the generation of a vast amount of data, particularly in the form of measurements and clinical notes. This data encompasses vital signs, laboratory test results, imaging reports, and physician notes, providing a comprehensive snapshot of the patient's health status \citep{johnson2016mimic}. The measurements serve as objective indicators of the patient's physiological state, while the clinical notes document healthcare professionals' observations, diagnoses, treatment plans, and other information. Collectively, this data plays a crucial role in providing valuable insights into a patient's condition, monitoring her/his progress, and guiding medical decisions throughout the ICU admission.

Effectively harnessing the abundance of data in the ICU poses significant challenges \citep{kruse2016challenges}, given the sheer volume of information, coupled with continuous monitoring and documentation, which can overwhelm healthcare providers. Extracting meaningful patterns and knowledge from this complex data requires sophisticated computational techniques and specialized algorithms. Furthermore, the data's diversity, heterogeneity, potential noise, and varying rates of missing values add extra layers of complexity to analysis and interpretation \citep{xiao2018opportunities}. In recent years, deep learning models have emerged as powerful tools for automatically learning tasks from large-scale data; however, they face a significant limitation due to their dependence on substantial labeled data for training \citep{fredriksson2020data}. Labeling data involves associating each data point with a specific target, such as a diagnosis or clinical outcome. In the ICU context, obtaining labeled data for training deep learning models is time-consuming, expensive, and often impractical \citep{xiao2018opportunities}. Expert annotations are required, which may not always be readily available due to privacy concerns, specialized expertise, and the urgency of medical decision-making in critical care settings. Modeling often relies on codes for labels that may not be available at the time of admission \citep{xiao2018opportunities}.

As a potential solution to this challenge, self-supervised pretraining has emerged. However, it's essential to note that several existing medical pretraining methods \citep{mcdermott2021comprehensive, 10.1145/3516367}, predominantly rely on a single modality. These unimodal approaches do not fully harness the potential of multiple modalities to enhance their performance through alignment strategies. This limitation represents a significant opportunity for further research and development in the field of medical AI. Furthermore, beyond self-supervised pretraining, multimodal pretraining methods \citep{radford2021learning, lu2019vilbert, li2019visualbert, su2019vl, zhang2020contrastive} have demonstrated the effectiveness of utilizing different modalities to pretrain models.

In the context of ICUs, clinical notes and measurements represent separate but complementary modalities that describe the same patient event. Clinical notes provide textual context, capturing subjective observations and interpretations from healthcare providers, while measurements offer objective numerical data reflecting the patient's physiological state. By incorporating both modalities during pretraining, deep learning models can potentially capture a more comprehensive understanding of the patient's condition and enable more accurate predictions.

To address the challenge of jointly embedding clinical notes and measurements from the same ICU stay, our devised training scheme enables multiple notes to be mapped to the embedding of a single ICU stay. This strategy allows notes and measurements to be aligned using JE-SSL methods. We enhance this pretraining process by incorporating masking, ultimately increasing the number of training examples, particularly benefiting the modality with the fewest available samples.

We proceed to evaluate the performance of a model pretrained through our method across in-hospital mortality (IHM) and phenotyping tasks. Additionally, we assess the ability of our pretraining to learn meaningful embeddings through linear classifier training and evaluate cross-modality retrieval. In the context of the IHM task, we introduce a zero-shot evaluation during pretraining. Lastly, we gauge our model's performance in a semi-supervised setting with limited labeled data.

Through these comprehensive evaluations, we aim to establish the efficacy of our proposed model in enhancing ICU data analysis, improving clinical decision-making, and unraveling complex critical care scenarios. Our contributions can be summarized as follows:

\begin{enumerate}
    \item We designed a novel pretraining approach that facilitates the alignment of clinical measurements and notes, enabling us to create a multi-modal pretraining objective.
    \item We evaluate the performance of our pretrained model across multiple tasks, demonstrating its ability to learn meaningful clinical representations and improve the performance of downstream tasks even with limited labeled data.
    \item We establish a zero-shot evaluation task for assessing the performance of multimodal pretrained models during the pretraining phase.
    
\end{enumerate}

\section{Related Works}
\label{sec:rel_works}

\subsection{Multimodal Pretraining}
 Multimodal pretraining methods have demonstrated success in preparing models for various downstream tasks \citep{radford2021learning, bardes2021vicreg}. In the following sections, we will introduce the current methods employed in multimodal pretraining.

\subsubsection{Contrastive Learning}

Contrastive learning methods create joint embeddings through similar and dissimilar views of data. For example, contrastive pretraining on image data \citep{chen2020simple} applies compositions of transformations to input images. Researchers have extended these techniques beyond a single modality, demonstrating that supervised language models help vision transformers learn semantic representations more effectively \citep{ghiasi2022vision}. Learning captions associated with images has also shown to improve visual models \citep{radford2021learning}. Other methods use both modalities together while separating their learned representations \citep{liang2022mind}. With medical data, one learning approach uses pairs of images and text, despite medical data lacking the high inter-class similarity found in other imaging applications \citep{zhang2020contrastive}. This work will use such contrastive approaches to align multimodal medical data. Similar to our method, \citet{raghu2022contrastive} develops a contrastive pretraining method for waveform and vitals measurements. However, our approach incorporates clinical notes instead of waveforms.

\subsubsection{Masked Pretraining}
Masked pretraining is a powerful technique that has demonstrated success in both language and vision domains. It involves training models to predict masked or missing portions of input data, effectively encouraging the model to learn meaningful representations of the entire input. This approach has been widely adopted in language models such as BERT \citep{devlin-etal-2019-bert}, where words are randomly masked in a sentence, and the model learns to predict the missing words based on the context of the surrounding words.

In the context of vision, masked pretraining has been applied to image data as well. Models like VisualBERT \citep{li2019visualbert} utilize masked image regions and captions to train joint vision-language representations. Similarly, in ViT \citep{he2022masked}, masked patches of images are used to learn representations that capture global visual context. This technique has enabled models to understand images beyond individual objects and focus on relationships between different regions.

Masked pretraining has also been extended to multimodal learning, such as in VL-BERT \citep{su2019vl}, where both text and image modalities are masked to predict missing parts, fostering a comprehensive understanding of their interactions. Furthermore, the idea of masked pretraining has been applied to healthcare settings, like in ClinicalBERT \citep{alsentzer2019publicly}, for learning representations from electronic health records, contributing to better medical data analysis.

\subsection{Medical Outcome Prediction}

In the realm of medical modeling, two common and critical task involve predicting in-hospital mortality and phenotyping, particularly in intensive care units (ICUs). The freely accessible MIMIC-III dataset has been extensively employed for such applications, utilizing biomedical text data \citep{ghassemi2014unfolding} as well as time-series data \citep{ghassemi2015multivariate, harutyunyan2019multitask}. Taking a step further, our work extends this by evaluating mortality and phenotyping prediction using a combination of both text and time-series tabular data. Notably, a fused model has demonstrated improved prediction performance in this context \citep{shukla2020integrating}. However, it's important to highlight that this fused model necessitates the presence of both modalities for making predictions. In contrast, our dual encoder setup offers the flexibility to utilize each modality individually after the pretraining stage, enhancing the adaptability of the model for diverse clinical scenarios.

\section{Methods}

\begin{figure*}[h]
    \centering
    \includegraphics[width=1\textwidth]{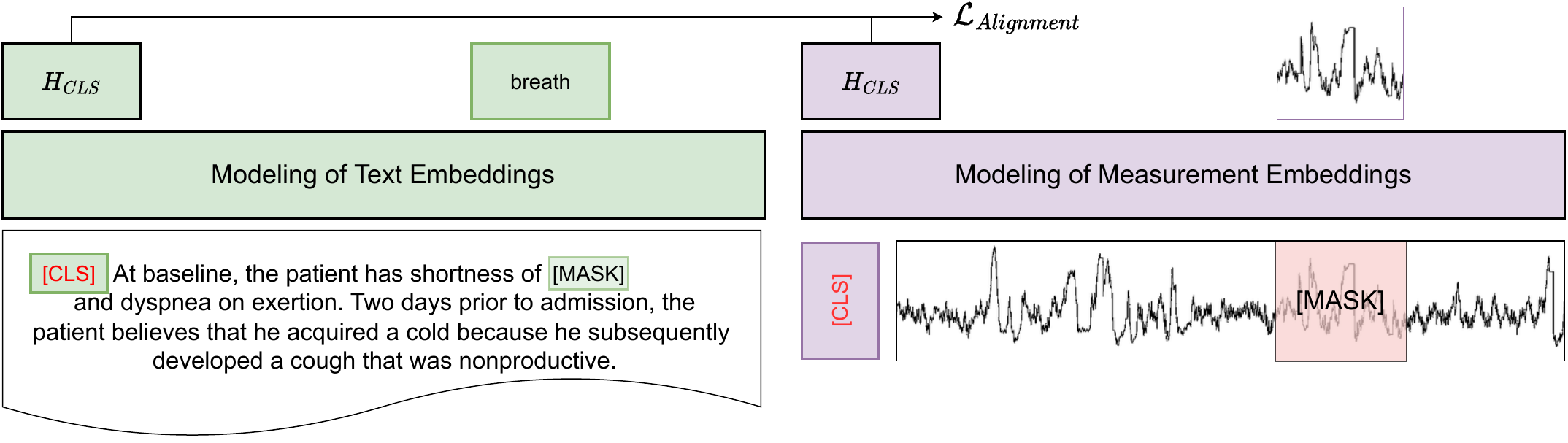}
    \caption{This figure provides a graphical representation of our proposed framework. Inputs from the different modalities are  transformed by their modality specific models. The two models are trained to align the class output of the final layer, $H_{CLS}$, and reconstruct masked tokens.}
    \label{fig:main}
\end{figure*}

Measurement time series and clinical notes from the same ICU stay are different modalities describing the same event, each containing additional information. However, the representation of text data in the form of token IDs and measurement time series varies drastically, making it challenging for machine learning models to discern the similarities between the two. To address this challenge, we aim to align these modalities within a joint embedding space. Our hypothesis is that the features learned for aligning these two modalities will also prove beneficial for downstream clinical tasks.

Our proposed method is comprised of two distinct phases. The first phase involves the pretraining of two modality specific encoders through alignment and masked prediction. Subsequently, in the second phase, we proceed to fine-tune each model separately on a range of downstream tasks. The objective function that we develop to pretrain our model can be broken down into two main component: a masked prediction component and a contrastive component. The following section describes the details of our proposed pretraining method. We provide a visual depiction of our pretraining in Figure \ref{fig:main}.

\subsection{Alignment Pretraining}

Throughout this section, our focus centers on a medical dataset, wherein each sample within the dataset consists of a timeseries of measurements, $\mathbf{M}$, and a set of medical notes, $\{ T_0 \cdots T_I \}$, where $I$ is the cardinality of the set. For each distinct sample, the collection of measurements and notes corresponds to a shared ICU stay. We introduce a measurement encoder, $E^M$, and text encoder, $E^T$, designed to process batches of sequences of measurements and notes, generating corresponding sequences of embedded representations, denoted as $\mathbf{H}^{M}$ and $\mathbf{H}^{T}$ for the measurements and a single clinical note respectively.

The core focus of our multi-modal training objective is on achieving alignment between two distinct modalities, guided by the principles of contrastive learning. In the context of JE-SSL \citep{radford2021learning}, the fundamental objective is twofold: to maximize the similarity among positive pairs of data while minimizing the similarity within negative pairs. For our context, positive pairs consist of measurements and notes attributed to the same ICU stay, while negative pairs consist of all other combinations of measurements and notes in the dataset. This involves mapping both measurements and notes into a joint embedding space, where the cosine similarity between their respective representations is maximized.

Integrating clinical notes into the learning process in clinical data presents challenges. Ideally, all ICU stay notes would be seamlessly incorporated, either through specialized models for extended sequences or consolidating them into a single representation. However, obstacles arise. Many ICU stays have over 100 notes, increasing computational demands for training. In addition, the effectiveness of contrastive training relies on ample positive pairs. However, in unimodal clinical contexts, the number of measurement windows is typically much smaller than the number of notes, which limits the availability of positive pairs.

Addressing these challenges, we introduce a method that efficiently manages computational resources while concurrently maximizing the abundance of training pairs, essential for effective contrastive pretraining. In our proposed method we consider every combination of measurements and notes. In doing so, we are able to increase the number of positive pairs to the total number of notes, $N$, in our dataset. We define our alignment objective as:

$$\mathcal{L}_{Alignment} = \mathcal{L}_{MT} + \mathcal{L}_{TM}$$

Where $\mathcal{L}_{MT}$ and $ \mathcal{L}_{TM}$ are defined as:

$$\mathcal{L}_{MT} = -\frac{1}{2N} \sum^N_{j=1}\log\underbrace{\left[\frac{\exp(\langle\mathbf{\hat{H}}_j^{M},\mathbf{\hat{H}}_j^{T}\rangle/\tau)}{\mathlarger{\sum}_{\mathbf{\hat{H}}^{-T} \in \mathcal{N}^T}\exp(\langle\mathbf{\hat{H}}_j^{M},\mathbf{\hat{H}}^{-T}\rangle/\tau)}\right]}_\text{Contrast measurements with notes}$$

$$\mathcal{L}_{TM} = -\frac{1}{2N}\sum^N_{k=1}\log\underbrace{\left[\frac{\exp(\langle\mathbf{\hat{H}}_k^{M},\mathbf{\hat{H}}_k^{T}\rangle/\tau)}{\mathlarger{\sum}_{\mathbf{\hat{H}}^{-M} \in \mathcal{N}^M}\exp(\langle\mathbf{\hat{H}}^{-M},\mathbf{\hat{H}}_k^{T}\rangle/\tau)}\right]}_\text{Contrast notes with measurements}$$

Here $\langle \cdot, \cdot\rangle$ is the inner product between the vectors, $\mathbf{\hat{H}}_j^{M}$ and $\mathbf{\hat{H}}_j^{T}$ is the normalized embedding of the class representation for the text and measurement modality respectively. $\tau$ is the temperature hyperparameter, $\mathcal{N}^T$ is the set of negative note pairs given a measurement window, and $\mathcal{N}^M$ is the set of negative measurement pairs given a note. The class representation of each modality is computed using a class token at the beginning of each sequence. For note embeddings, this is done using a reserved class embedding at the beginning of each sequence. For measurements, we utilize a learned class token at the beginning of each sequence similar to \citet{dosovitskiy2020image}. By minimizing this loss, the similarity between positive embedding pairs is increased while the similarity between negative pairs is decreased. 

In many contrastive training objectives, negative pairs are considered to be all other pairs in a dataset, However, in our setting, it does not make sense to decrease the similarity between a pair of measurements and notes from the same ICU admission. To avoid this situation, we consider the set of negative pairs to be samples from other ICU stays. In practice, we train our models on this objective by randomly sampling a note from each ICU stay as a positive pair during each epoch.

\subsection{Masked Pretraining}

Incorporating insights from recent advancements like ViL-BERT \citep{lu2019vilbert}, which emphasize combining alignment prediction and masked token prediction in multi-modal pretraining, we adopt a similar approach. We tokenize clinical notes using the ClinicalBERT \citep{alsentzer2019publicly} tokenizer, originally trained on the MIMIC-III dataset's notes. Measurement data is embedded at each time step in our sequence using a linear projection of the input data. Our masking strategy replaces notes with a mask embedding token and measurements with the mean measurement values.

Following the principles of \citet{he2022masked, devlin-etal-2019-bert}, our method involves forwarding the final layer's output through a linear layer. This linear layer adjusts the output's dimensionality for the subsequent reconstruction task: for notes, it aligns with the number of note tokens, and for measurements, it corresponds to the size of a measurement at a specific time. The reconstruction loss focuses solely on the masked tokens and their reconstructed counterparts.

For notes, we employ cross-entropy loss to measure the difference between predicted target tokens and actual token values. For measurements, we use the Smoothed L1 loss \citep{girshick2015fast} to prevent exploding gradients for very large or small measurement values. Our final objective combines the contributions of the masked and contrastive objectives.

\subsection{Model Architectures}

In this section, we define the architectures used during our experiments. We choose to use transformer based models for both our modalities. More details about their implementations are in the following sections.

\subsubsection{Text Encoder Architecture}

Our text encoder is based on the BERT architecture. The input to the model consists of tokenized medical notes. These tokens are mapped to corresponding learnable embeddings. To account for the sequential order of tokens, we add a positional embedding to the token embeddings. The resultant embeddings are then passed through a multi-layer transformer model, ultimately generating the final embedding representation for the text modality. A linear layer is used to project the last hidden state for token prediction. An additional linear layer is used for the last hidden state of the class token so that it can be projected to the same size as the measurement encoders class representation.

\subsubsection{Measurement Encoder Architecture}

The architecture of our measurement encoder is akin to that of the text encoder, with a distinction in the initial embedding process. We opt to employ a linear layer to embed the measurement data into token embeddings. In addition, a fixed sinusoidal positional embedding is combined with the token embeddings. The resulting embeddings are subsequently fed through a transformer model architecture to produce the final embedding for the measurement modality. Similar to the text encoder, linear layers are used to project the final layer for masked reconstruction and class token alignment.

\section{Experiments}

In this section, we outline the experimental framework designed to assess the effectiveness of our model across both pretraining and downstream tasks. The subsequent subsections detail the dataset employed for experimentation, encompassing any preprocessing procedures undertaken. We then proceed to elaborate on our pretraining experiments and describe the metrics employed for evaluating our model's performance in this phase. Finally, we describe the methodology applied to train and evaluate our model on various downstream tasks. Our implementation is written using the PyTorch library \citep{paszke2019pytorch} using 1 NVIDIA RTX A5000 GPUs. The code for our experiments is publicly available on GitHub: \href{https://github.com/kingrc15/multimodal-clinical-pretraining}{https://github.com/kingrc15/multimodal-clinical-pretraining}

\subsection{Data}

In our experimental setup, we leverage the publicly accessible MIMIC-III dataset, a comprehensive repository of de-identified health records from patients admitted to intensive care units (ICUs). Within this dataset, a wealth of clinical measurements, medical notes, and related data is available, all associated with unique hospital admission stay identifiers. For the measurement data, we apply the preprocessing pipeline used in \citet{harutyunyan2019multitask} resulting in uniform sequences of 17 measurements. We utilize the same train, validation, and test splits used in the benchmark. 

Simultaneously, we construct the notes dataset by merging various types of notes with measurements, emphasizing information relevant to our downstream tasks. This process significantly augments the measurement dataset, enhancing the potential for learning meaningful representations. 

During fine-tuning, we use the MIMIC-III benchmark dataset \citep{harutyunyan2019multitask}, maintaining the same train, validation, and test splits while incorporating additional hospital admissions that were excluded during the initial data preparation. For a full description of the data, see Appendix \ref{apd:data}.

\subsection{Pretraining}

We train the described text and measurement encoders using our proposed pretraining objective. Each of the modality specific architectures is a transformer consisting of 8 layers with a hidden dimension of 128, 8 attention heads, and a GELU activation function. We train our model with an AdamW optimizer \citep{loshchilov2017decoupled} with a cosine annealing learning rate schedule \citep{loshchilov2017sgdr}. We use the train split of our multi-modal dataset to train our models while the validation dataset is used for tuning hyperparameters.

Building upon previous research in multimodal pretraining \citep{radford2021learning}, we perform an evaluation that centers on the model's capacity to retrieve relevant samples from the complementary modality. This evaluation entails measuring the similarity between pairs of samples within both the measurement and notes datasets. To be specific, we calculate the recall at the top-k retrieval, where k represents the number of retrieved items. For the purpose of tuning our model's hyperparameters, we utilize the mean of the top-1 recall for both modalities on our held-out validation dataset. The results, including recall values at the top-1, 5, and 10, for our best-performing model pretrained with alignment only and alignment + masking, are presented in Table \ref{tab:recall}. 

We used the best hyperparameters from our grid search and retrained a model for 100 epochs. This model was then used for our downstream tasks.

\subsection{Downstream Tasks}

Our objective is to determine if pretraining is useful for learning essential features for downstream tasks. To assess this, we conducted evaluations on two downstream tasks. The first is the IHM benchmark task. This task involves binary classification, wherein the model predicts whether a patient will pass away during their ICU stay, based on the initial 48 hours of available information. To evaluate performance, we adopted the area under the curve of the receiver operating characteristic curve (AUC-ROC) and area under the precision recall curve (AUC-PR) metrics to measure the alignment between the model's predicted labels and the true labels.

Our second task is the phenotyping benchmark, which is a multi-class multi-label classification task. In this scenario, the model's objective is to predict the presence of any of the 25 given phenotypes, utilizing the measurements collected during an ICU stay. For assessing model performance, we employed the macro-AUC-ROC and micro-AUC-ROC metrics, facilitating an evaluation of the alignment between the model's predicted labels and the true labels.

For both of these tasks, we employ the measurement encoder with a linear classifier initialized with random weights. This linear layer, transforming the class token embedding from the final layers into predicted labels. An AdamW optimizer \citep{loshchilov2017decoupled} is used to update the model along with a cosine annealing learning rate schedule \citep{loshchilov2017sgdr}. 

In the following sections we describe three experiments that evaluate the ability of our pretraining to learn meaningful representations for downstream tasks. In each of the settings, our best pretrained model is used and compared to a randomly initialized model along with related methods. 

\subsection{Zero-Shot Evaluation}

We introduce a zero-shot evaluation method using the IHM task. Our approach initiates by processing two distinct phrases through our text encoder: "patient deceased" and "discharged today." These phrases serve as indicators of whether a patient expired during their ICU stay or was discharged. Subsequently, for each measurement in our held out test dataset, we apply our measurement encoder. We extract the class token representations from the final layer of each model and compute the cosine similarity between each pair. These cosine similarity scores are then transformed into a probability distribution through a softmax function. These probabilities are treated as the model's predicted labels, which are subsequently compared with the ground truth labels for evaluation.

Intuitively, this evaluation measures the alignment of the measurement data with our two anchor phrases and assigns a probability score that indicates their closest connection. This process allows us to effectively assess the degree of alignment between the embedding space and the text, all without the need for explicit labels. It's a tailored approach, specifically designed for the IHM task.

\subsection{Linear Evaluation}

Following previous work on JE-SSL pretraining \citep{chen2020simple}, we follow the linear evaluation protocol where a linear classifier is trained on top of the frozen pretrained model. The objective of this evaluation is to understand how well the features learned by the JE-SSL model can generalize to new, task-specific data. If the features are generalizable, the linear classifier should achieve reasonable performance on the downstream task, even without fine-tuning the whole model.

\subsection{Semi-Supervised Evaluation}

We extend our analysis by conducting supplementary experiments aimed at understanding the influence of the pretraining phase on downstream classification tasks with minimal labeled data. The hypothesis is that a well-pretrained model requires fewer labeled examples during the fine-tuning process. To investigate this, we conduct three distinct semi-supervised experiments, using 1\%, 10\%, and 50\% of the available data as training samples.  We subsequently assess the model's performance on withheld datasets to ascertain its ability to generalize under constrained labeled data scenarios. This approach affords insights into the pretraining's efficacy concerning data efficiency and the model's adaptability to scenarios with minimal labeled data.

\section{Results and Discussion}

\begin{table*}[h!]
\begin{center}
\begin{tabular}{cccccc}
                          &        & \multicolumn{2}{c}{In-Hospital-Mortality}   & \multicolumn{2}{c}{Phenotyping} \\ \hline
Modality                  & Labels & AUC-ROC       & AUC-PR        & Macro AUC-ROC & Micro AUC-ROC \\ \hline
LR$^*$                    & 100\%  & 0.848 (0.020) & 0.301 (0.001) & 0.739 (0.005) & 0.799 (0.003) \\ \hline
\multirow{2}{*}{LSTM$^*$} & 50\%   & 0.836 (0.004)  & 0.448 (0.011) & 0.746 (0.002) & 0.801 (0.002) \\ 
                          & 100\%  & 0.855 (0.020) & 0.485 (0.053) & 0.770 (0.004) & 0.821 (0.003) \\ \hline
\multirow{3}{*}{Baseline} & 1\%    & 0.528 (0.097) & 0.140 (0.036) & 0.563 (0.004) & 0.702 (0.003) \\
                          & 10\%   & 0.794 (0.008) & 0.373 (0.008) & 0.655 (0.018) & 0.740 (0.007) \\
                          & 50\%   & 0.831 (0.004) & 0.455 (0.014) & 0.727 (0.006) & 0.787 (0.006) \\ \hline
\multirow{4}{*}{Ours}     & 0\%    & 0.709 (0.000) & 0.214 (0.000) & - & - \\
                          & 1\%    & 0.706 (0.024) & 0.237 (0.017) & 0.659 (0.002) & 0.743 (0.003) \\
                          & 10\%   & 0.815 (0.010) & 0.390 (0.012) & 0.703 (0.004) & 0.766 (0.006) \\
                          & 50\%   & 0.850 (0.003) & 0.493 (0.004) & 0.727 (0.001) & 0.788 (0.001) \\ 
                          & 100\%  & 0.856 (0.004) & 0.495 (0.005) & 0.742 (0.003) & 0.798 (0.003) \\ \hline
\end{tabular}
\end{center}
\caption{We present the results of our semi-superivsed experiments, both with and without pretraining. We report the mean results and standard deviations obtained from five runs. * results are from \citet{harutyunyan2019multitask}}
\label{tab:semi_eval}
\vspace{-10pt}

\end{table*}

\begin{table}[]
\begin{tabular}{cccc}
                                                & Method   & AUC-ROC             & AUC-PR        \\ \hline
\multirow{2}{*}{\rotatebox[origin=c]{90}{IHM}}  & Baseline & 0.855 (0.020)       & 0.485 (0.053) \\
                                                & Ours     & 0.828 (0.001)       & 0.402 (0.001) \\ \hline
                                                &          & Macro               & Micro         \\ \hline
\multirow{2}{*}{\rotatebox[origin=c]{90}{Phen}} & Baseline & 0.770 (0.004)       & 0.821 (0.003) \\
                                                & Ours     & 0.710 (0.001)       & 0.776 (0.001) \\
\end{tabular}

\caption{We present the results of our linear evaluation experiments for IHM and Phenotyping (Phen), with the MIMIC-III Benchmark's fully supervised model as a reference. We report the mean results and standard deviation from five runs. }
\label{tab:lin_eval}
\vspace{-20pt}
\end{table}

In this section we report the results of our evaluation experiments. We report the zero-shot, linear and the semi-supervised evaluation results and discuss below.

\subsection{Zero-Shot Evaluation}

Using Subsequently, we evaluate this pretrained model on the test split, where it attains an AUC-ROC of 0.709 and an AUC-PR of 0.214. This pretrained model serves as the foundation for our subsequent evaluations. Appendix \ref{apd:grid} contains a detailed list of the best hyperparameters resulting from this search.

Remarkably, even in the absence of labeled data, our measurement model demonstrates the capacity to acquire features that hold significance for IHM tasks. Although these results are notably below fully supervised scores, this evaluation provides valuable insights into the learning process of the measurement model during pretraining.

\subsection{Linear Evaluation}

We adopt a grid search approach to optimize hyperparameters for this task, specifically weight decay, learning rate, and the number of training epochs. Our hyperparameter selection is based on minimizing the loss on the validation dataset after training on the training dataset. The search space for hyperparameters is comprehensively outlined in Appendix \ref{apd:grid}.

In Table \ref{tab:lin_eval}, we present the results on our test dataset, comparing them with a fully supervised model from the MIMIC-III benchmark for reference. Notably, our findings indicate that a linear classifier trained on the fixed representations of the measurement model achieves performance levels nearly equivalent to those of a fully supervised model. This observation underscores the substantial utility of the learned representations for both phenotyping and in-hospital mortality prediction tasks and demonstrates that the features learned for alignment are also useful for downstream tasks.

\subsection{Semi-Supervised Evaluation}

We conduct a grid search for hyperparameters using the train and validation dataset. For more details, see Appendix \ref{apd:grid}.

Table \ref{tab:semi_eval} presents our results on the test dataset. In this evaluation, we compare our method with two models trained on all the data from the MIMIC-III benchmark, which serve as reference points. Additionally, we assess our method against a randomly initialized model trained on the same data, providing a baseline for evaluating the effectiveness of our pretraining.

Our results indicate that our proposed pretraining scheme performs optimally when the percentage of labeled data is smaller. We observe the most significant improvements when only 1\% of the labels are available for both IHM and Phenotyping tasks in comparison to a randomly initialized model. Furthermore, our pretrained model achieves nearly identical results to a fully supervised model when only half of the labels are available for the IHM task. This demonstrates the efficacy of our pretraining approach, particularly in semi-supervised scenarios. 

Additionally, we observed an improvement in phenotyping results when 1\% and 10\% of the labels were available. However, this increase was not observed with a higher percentage of labels, specifically for the phenotyping task. At 50\% of the labels our approach outperforms the LSTM-based baseline for the IHM task (0.850 AUC-ROC versus 0.835, with non-overlapping confidence intervals). The results for the phenotyping task are similar, with our model achieving a micro score of 0.79 versus the benchmark model of 0.80. At 50\%, our model performs similarly to the randomly initialized models. While the linear evaluation results lead us to believe the pretrained model is learning useful features for this task, the reason for this decline in performance is unclear. We believe that our model may be suffering from catastrophic forgetting, a phenomenon where learned features are forgotten \citep{french1999catastrophic}. Additionally, due to the sparsity of some of the labels, and their clinical similarity, it is possible that the model is forgetting key features that would differentiate the phenotypes from each other. This issue may be addressed through catastrophic forgetting mitigation methods as well as further clinical interpretation of the learned representations, and we leave it as a key limitation to be addressed in future work.

\section{Limitations and Conclusions}

The primary focus of our work is to develop a pretraining method that enriches a measurement model with insights from clinical notes. We have demonstrated that our pretraining method is effective in learning valuable representations for various clinical prediction tasks across data modalities. However, it's important to acknowledge that our proposed training scheme does place limitations on the expressive capabilities of the notes model. Future research avenues could concentrate on techniques that simultaneously develop both text and measurement encoders. 

Furthermore, while we have introduced the IHM zero-shot evaluation as one criterion for assessing the measurement encoder during pretraining, it's worth noting that other zero-shot tasks for various benchmark tasks could be devised. However, determining the most effective evaluation approach across all tasks remains an open question.

Notably, our pretraining method and model excels in scenarios where labeled data is scarce. This is demonstrated by our semi-supervised experiments, which utilize only a small fraction of the available labels and achieve significant improvements in both AUC-ROC and AUC-PR metrics. This has the potential to significantly reduce the labeling costs for clinics when training models with data from their EHRs and improve accuracy of models deployed in these settings.

\bibliography{king23}

\appendix

\section{Retrieval Task}\label{apd:ret}

The results, including recall values at the top-1, 5, and 10, for our best-performing model pretrained with alignment only and alignment + masking, are presented in Table \ref{tab:recall}. We found that recall for both setting remain very small. However, we do note that the addition of masking increased the recall significantly.

\begin{table*}[]
\begin{center}
    
\begin{tabular}{ccccccc}
           & \multicolumn{3}{c}{M2N} & \multicolumn{3}{c}{N2M} \\ \hline
           & R@1    & R@5    & R@10  & R@1    & R@5    & R@10  \\ \hline
Align      & 0.065  & 0.195  & 0.391 & 0.261  & 0.977  & 1.498 \\ \hline
Align+Mask & 0.758  & 4.090  & 7.123 & 1.011  & 4.000  & 7.330  \\ \hline
\end{tabular}
\caption{We present the recall values for our top-performing models, both with and without masking, for two tasks: M2N (notes given measurement retrieval) and N2M (measurements given notes retrieval)}
\label{tab:recall}
\end{center}

\end{table*}

\section{Hyperparameter Search}\label{apd:grid}

To select the hyperparameters for our pretraining, we conduct a grid search over 5 epochs for learning rate, weight decay, measurement and notes dropout rates, the temperature, the batch size, and mask rate for each modality. For learning rate we used 0.0001, 0.00001, and 0.000001. For weight decay we used 0.2, 0.1, and 0.01. For measurement and note dropout we used 0.0 and 0.1. For temperature we used 0.1, 0.07, and 0.05. For batch size we used 16, 32, 64, and 128. For measurement and note masking we used 0.0, 0.1, and 0.2.

To select the hyperparameters for linear evaluation, we conduct a grid search over batch size, epochs, learning rate, and weight decay. For batch size, we used 8, 16, 32, and 64. For epochs, we used 1, 2, 3, 4, and 5. For learning rate, we used 1e-3, 1e-4, and 1e-5.

To select the hyperparameters for our semi-supervised experiments, we conduct a grid search over batch size, epochs, learning rate, and weight decay. The search space for batch size, learning rate and weight decay are the same as the linear evaluation. For the pretrained model, the epochs are the same as the linear evaluation. For fair comparison, we allow the randomly initialized model to train longer as we notice that it performs better with more epochs. The epochs used during the search for the randomly initialized model were 5, 10, 15, and 20. 

Following \citet{chen2020simple}, we also use a different higher learning rate for the randomly initialized linear classifier when finetuing our pretrained model. This is done to avoid catastrophic forgetting. The search space for the linear classifier learning rate is 1e-2, 1e-3, and 1e-4.

\section{Data Description}\label{apd:data}

Our experimental setup involves the utilization of the MIMIC-III dataset \citep{johnson2016mimic}, a publicly accessible critical care database known as the Medical Information Mart for Intensive Care III (MIMIC-III). This dataset contains de-identified health records of patients admitted to intensive care units (ICUs). Within MIMIC-III, a diverse range of clinical measurements, medical notes, and related information is available, each linked to unique identifiers for ICU stays. These identifiers enable us to match measurements and notes corresponding to the same ICU stay, thus forming a focused subset for our analysis.

To construct this subset, we start by forming the measurement subset. Following the methodology outlined in \citet{harutyunyan2019multitask}, we acquire our measurement data. Specifically, we select 17 clinical features encompassing both categorical and continuous values from each ICU stay. We exclude cases involving ICU transfers, multiple ICU stays, pediatric patients, and instances with invalid or missing IDs. A series of measurements is then created using uniform time intervals, with missing values imputed based on previous measurements. In cases where preceding measurements are unavailable, normal values are imputed. Subsequently, the features are normalized before being fed into the measurement encoder. We split this dataset into training and testing subsets, with the test set comprising 15\% of the split. Additionally, we further partition the training data into training and validation subsets, maintaining these same splits during both the pretraining and fine-tuning phases. For a comprehensive description of the preprocessing pipeline, refer to \citet{harutyunyan2019multitask}.

Subsequently, we proceed to establish the notes datasets by utilizing the MIMIC-III notes table. The dataset comprises various types of notes, each of which can be matched with measurement windows. To focus on pertinent information for our downstream tasks, certain measurements, such as "Social Work" and "Case Management," are excluded. In our dataset, we retain the following note types: Echo, ECG, Nursing, Physician, Respiratory, Radiology, and Discharge Summary. We merge this table of notes to measurements based on their Hospital Admission ID, using an inner join. We filter out notes that do not happen within the window of the measurements for all notes except for Discharge Summaries which provide a summary of the Hospital Admission. Finally, we set a max sequence length of 256 for both modalities. For windows that are longer than the predefined window size, we select a random window of measurements of the max size during each training iteration.

This process of matching notes with measurements significantly augments the measurement dataset, expanding it from an initial count of 24,864 to a total of 186,357 positive pairs. This expansion contributes an increased number of representations for the measurement encoder to learn from, enhancing its training potential. During fine-tuning, we simply use the MIMIC-III benchmark dataset which contains the same train, validation, and test splits but contains more hospital admissions that were excluded during the join of the measurement and notes.

\end{document}